\documentclass{article}

\usepackage[final]{AdvML_Frontiers_2024} 

\usepackage[utf8]{inputenc} 
\usepackage[T1]{fontenc}    
\usepackage{hyperref}       
\usepackage{url}            
\usepackage{booktabs}       
\usepackage{amsfonts}       
\usepackage{nicefrac}       
\usepackage{microtype}      
\usepackage{xcolor}         

\usepackage{graphicx}
\usepackage{subcaption}
\usepackage{amsmath}
\usepackage{multirow}
\usepackage{cleveref}

\title{Seeing Through the Mask: \\ Rethinking Adversarial Examples for CAPTCHAs}

\author{%
  Yahya~Jabary\\
  TU Wien\\
  \texttt{jabaryyahya@gmail.com} \\
  \And
  Andreas~Plesner\thanks{Corresponding author} \\
  ETH Zurich \\
  \texttt{aplesner@ethz.ch}
  \AND
  Turlan~Kuzhagaliyev \\
  ETH Zurich \\
  \texttt{kturlan@student.ethz.ch} 
  \And
  Roger~Wattenhofer \\
  ETH Zurich \\
  \texttt{wattenhofer@ethz.ch} \\
}

\begin{document}

\maketitle

\begin{abstract}

    

    Modern CAPTCHAs rely heavily on vision tasks that are supposedly hard for computers but easy for humans. However, advances in image recognition models pose a significant threat to such CAPTCHAs. These models can easily be fooled by generating some well-hidden "random" noise and adding it to the image, or hiding objects in the image. However, these methods are model-specific and thus can not aid CAPTCHAs in fooling all models. We show in this work that by allowing for more significant changes to the images while preserving the semantic information and keeping it solvable by humans, we can fool many state-of-the-art models. Specifically, we demonstrate that by adding masks of various intensities the Accuracy @ 1 (Acc@1) drops by more than 50\%-points for all models, and supposedly robust models such as vision transformers see an Acc@1 drop of 80\%-points. 
    These masks can therefore effectively fool modern image classifiers, thus showing that machines have not caught up with humans -- yet. 
\end{abstract}

\section{Introduction}


Not surprisingly, CAPTCHAs are currently threatened by advanced image recognition models. \citet{20.500.11850/679226} has recently shown that the most popular CAPTCHA environment (reCAPTCHA by Google~\citep{captchashare}) can be solved equally well by machines and humans. If CAPTCHAs are to have a future, a new approach is needed. Adversarial machine learning is closely related to CAPTCHAs, as researchers try to build samples where the machine fails to recognize the image while the human does not register any manipulation happening. On the one hand, these imperceptible manipulations are more ambitious than CAPTCHAs since even the earliest CAPTCHAs did not bother to hide the manipulation of the input. On the other hand, adversarial image generation is not robust enough for automatic bot detection, as it often tailors the attack to a specific model.
% ...
We want images that can effectively fool any machine learning model, but we do not mind having a visible manipulation. However, the manipulation should be easy for humans to filter out. In other words, we do not mind if many pixels are changed a lot, as long as the image is still easily recognizable to humans. This is easily achieved if the image manipulation is somehow predictable, for instance by overlaying the original with a periodic signal like a grid. A promising new form of CAPTCHAs, known as hCaptcha, is doing exactly that, and in this work, we want to get a clearer understanding of what this approach can and cannot do.

The signals, or masks, inspired by hCaptcha can be surprisingly simple yet very powerful. In addition, to fully assess their capabilities and potential impact on vision models, we have established the following key motivations for this study.

\begin{enumerate}
    \item \textbf{Exploration of aggressive adversarial perturbations}: In contrast to traditional adversarial attacks that aim for imperceptibility, our study focuses on the domain of CAPTCHAs where visible perturbations are acceptable. In this context, we can be more aggressive with the perturbations, as the limit is not imperceptibility but rather semantic preservation for humans.

    \item \textbf{Exploiting the human-machine vision gap}: Our research aims to highlight and leverage the difference in human and machine perception. 

    \item \textbf{Accessibility of attacks}: The simplicity and ease of execution of the proposed attacks make them readily available to large-scale CAPTCHA systems. 

    \item \textbf{Evaluating robustified models}: We aim to benchmark models that have been specifically fine-tuned for robustness in our use case. This evaluation will provide valuable insights into the effectiveness of current robustification techniques against our proposed class of adversarial examples.

\end{enumerate} 


Thus, our work examines adversarial examples through the lens of CAPTCHA services. We challenge the constraints of imperceptibility in adversarial attacks, proposing that any semantics-preserving distortion that effectively differentiates human users from automated solvers is acceptable within this domain. This approach allows for large perturbations, shifting our focus to metrics that quantify semantic change rather than visual imperceptibility.

Although reCAPTCHA has been broken, hCaptcha remains undefeated in the ongoing attack-defense arms race and has recently added multiple new challenges and layers of security measures~\citep{qin2dim,qin2dim2023challenge}.

\paragraph{Approach}

To investigate these issues, we focus on evaluating the performance of state-of-the-art vision models against a range of image filters inspired by hCaptcha techniques. Our study aims to:

\begin{enumerate}
    \item Quantify the drop in Acc@1 and Acc@5 accuracy when various filters are applied to input images.
    \item Compare the resilience of different model architectures to these adversarial examples.
    \item Assess whether models specifically designed for robustness offer significant advantages in this context.
\end{enumerate}

Our preliminary findings underscore the effectiveness of masks in challenging even the most advanced vision models, motivating our deeper investigation of these adversarial techniques.

Through this research, we hope to contribute to the ongoing discussion on AI safety and reliability, emphasizing the need for vision models that can maintain high performance in the face of real-world image manipulations. Our findings have implications not only for the development of more robust models but also for the broader challenge of creating computer vision systems that can match human-level adaptability in visual perception tasks.



\section{Related Work}

Deep learning models have achieved unprecedented performance in computer vision tasks, frequently exceeding human-level accuracy on image classification benchmarks~\citep{He_2015_ICCV,russakovsky2015imagenet}. State-of-the-art architectures such as Vision Transformers (ViT)~\citep{dosovitskiy2021imageworth16x16words}, ConvNeXt~\citep{Liu_2022_CVPR}, and EVA-02~\citep{fang2024eva} now form the foundation of numerous critical applications, ranging from autonomous vehicles~\citep{9046805} to medical imaging~\citep{chen2022recent, shamshad2023transformers}. However, the robustness of these models against adversarial attacks remains a pressing concern for their deployment in real-world scenarios, which could compromise their reliability and security~\citep{serban2020adversarial}.

The field of adversarial examples in machine learning has seen significant advances in recent years~\citep{Hendrycks_2021_CVPR}. Our work on geometric masks for CAPTCHAs builds on the foundational concept of robust and non-robust features in machine learning models, as proposed by ~\citep{ilyas2019adversarial}. This perspective suggests that adversarial examples exploit non-robust features susceptible to imperceptible perturbations while preserving robust features crucial for human interpretation.

Expanding on this framework, recent studies have demonstrated the potential of geometric metrics to detect adversarial samples. \citet{venkatesh2022detecting} showed promising results using density and coverage metrics to identify adversarial examples in datasets such as MNIST and biomedical imagery. This approach aligns with our focus on geometric perturbations that disrupt machine learning models' reliance on non-robust features while maintaining image semantic integrity for human solvers.



In the specific context of CAPTCHAs, researchers have explored various innovative approaches to enhance security against automated solvers. \citet{sheikh2022novel} proposed a novel animated CAPTCHA technique based on the persistence of vision, which displays text characters in multiple layers within an animated image. This word-level adversarial attack demonstrates ongoing efforts to develop more robust CAPTCHA systems that can effectively distinguish between human and machine solvers. Similarly, \citet{Hajjdiab2017RandomIM} introduced a random CAPTCHA system to match images that eliminates the need for an image database while maintaining ease of use. Their approach generates random images and asks users to match feature points between two images, leveraging concepts from computer vision research.

By synthesizing these diverse research directions, our work aims to contribute to the ongoing efforts to enhance the robustness of machine learning models against adversarial attacks, particularly in the context of CAPTCHA systems. We seek to leverage insights from geometric perturbations, adversarial training, and innovative CAPTCHA designs to develop more effective and secure visual challenges that maintain a clear distinction between human and machine solvers.

\section{Methodology}

In this section, we will go over the data that we used for the analysis along with the model choices. We have selected multiple models, which we will evaluate on the datasets to demonstrate the effectiveness of the masks we have constructed.

\paragraph{Models}

We selected several models to evaluate the performance of, namely: ``ConvNeXt\_XXLarge''~\citep{Liu_2022_CVPR}, Open CLIP's ``EVA01-g-14-plus''~\citep{Fang_2023_CVPR} and ``EVA02-L-14''~\citep{fang2024eva}, ``DFN5B-CLIP-ViT-H'' by Apple \citep{fang2023data}, the original ``ViT-L-14-378'' and ``ViT-H-14-378-quickgelu''~\citep{dosovitskiy2021imageworth16x16words}, ``ResNet50x64''~\citep{He_2015_ICCV}, and RoBERTa-B and RoBERTa-L~\citep{conneau2020unsupervisedcrosslingualrepresentationlearning}; the RoBERTa models are selected as they are supposed to be robust against adversarial attacks.\footnote{We highlight results for a subset of these, namely ConvNeXt, EVA02, ViT-H-14, ResNet50, and RoBERTa-L, and leave the rest for the appendix.} Due to time constraints, we were not able to test the method presented recently by~\citet{fort2024ensembleeverywheremultiscaleaggregation}; we leave this for future work. The models were selected to represent landmark architectures in both convolutional and transformer-based approaches. This selection allows us to evaluate the effectiveness of our masks across different model paradigms. 

\begin{figure}[t]
    \input{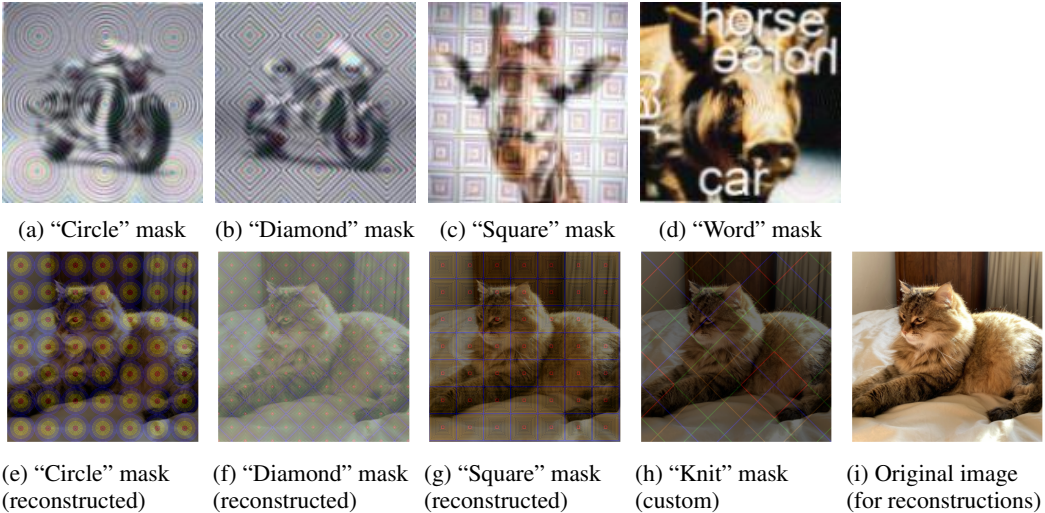}
    \caption{Selected examples by hCaptcha and their optimized reconstructions. The ``Word'' overlay was omitted and replaced with a custom ``Knit'' mask.}
    \label{fig:hcaptcha-combined}
\end{figure}

\paragraph{Data}

We conducted our experiments using both the enriched ImageNet dataset with 1,000 entries provided by ``visual-layer'' on HuggingFace and the reduced ImageNette dataset~\citep{Howard_Imagenette_2019}. The ImageNette dataset, consisting of approximately 10,000 images evenly distributed across 10 categories, was chosen to make the computations more feasible. To accommodate the need for multiple iterations on each image, we created three smaller datasets: \texttt{SubSet200}, \texttt{SubSet500}, and \texttt{ResizedAll}. \texttt{SubSet200} and \texttt{SubSet500} contain 2,000 and 5,000 images, respectively, maintaining the full resolution of ImageNette. \texttt{ResizedAll} includes all ImageNette images scaled down to 128x128 pixels, a standard size for CAPTCHAs, to speed up image processing. Note that this resizing may result in a slight performance drop compared to full-resolution images. The models generally achieve Acc@1 accuracy in the high 80\% to low 90\% range, with Acc@5 accuracy in the high 90\% range; see \Cref{sec:Acc@1 and Acc@5 All Clean} for details.

We defined four masks -- ``Circle'', ``Diamond'', ``Square'' and ``Knit'' -- which we apply to the images at various intensities. These masks were selected based on an experiment involving 1,600 web-scraped and hand-labeled images from hCaptcha. The number and intensity of mask elements are determined by the density and opacity values, with the density fixed to a constant value in our subsequent experiments focusing on the effects of varying opacity; for details, see \Cref{sec:hyperparam}.

\paragraph{Perceptual Quality and the Accuracy Metric}

Perceptual quality is a crucial aspect of our evaluation, assessing the visual fidelity of adversarial examples. We used a weighted average metric to capture various aspects of image quality. This metric combines cosine similarity (15\% weight)~\citep{singhal2001modern}, Peak Signal-to-Noise Ratio (PSNR, 25\% weight)~\citep{9311108}, Structural Similarity Index (SSIM, 35\% weight)~\citep{wang2004image}, and Learned Perceptual Image Patch Similarity (LPIPS, 25\% weight)~\citep{lpips}. The weights were chosen to balance the importance of each component in the overall quality assessment.

Moreover, we evaluate the models based on their accuracy. The models predict a likelihood for each of their pre-trained classes so the classes can be sorted by likelihood in descending order from top to bottom. We focus on the accuracy@k (with $k=1$ and $k=5$), denoted Acc@$k$, which measures how often the ground truth label is in the top $k$ classes.

\section{Results}

We perform three experiments, one per dataset, with the range of models mentioned earlier. We only show the key partial results here with full tables in the Appendix.

\subsection{Experiment 1 -- \texttt{SubSet500}}

\begin{figure}[t]
  \centering
  \includegraphics[width=0.9\linewidth]{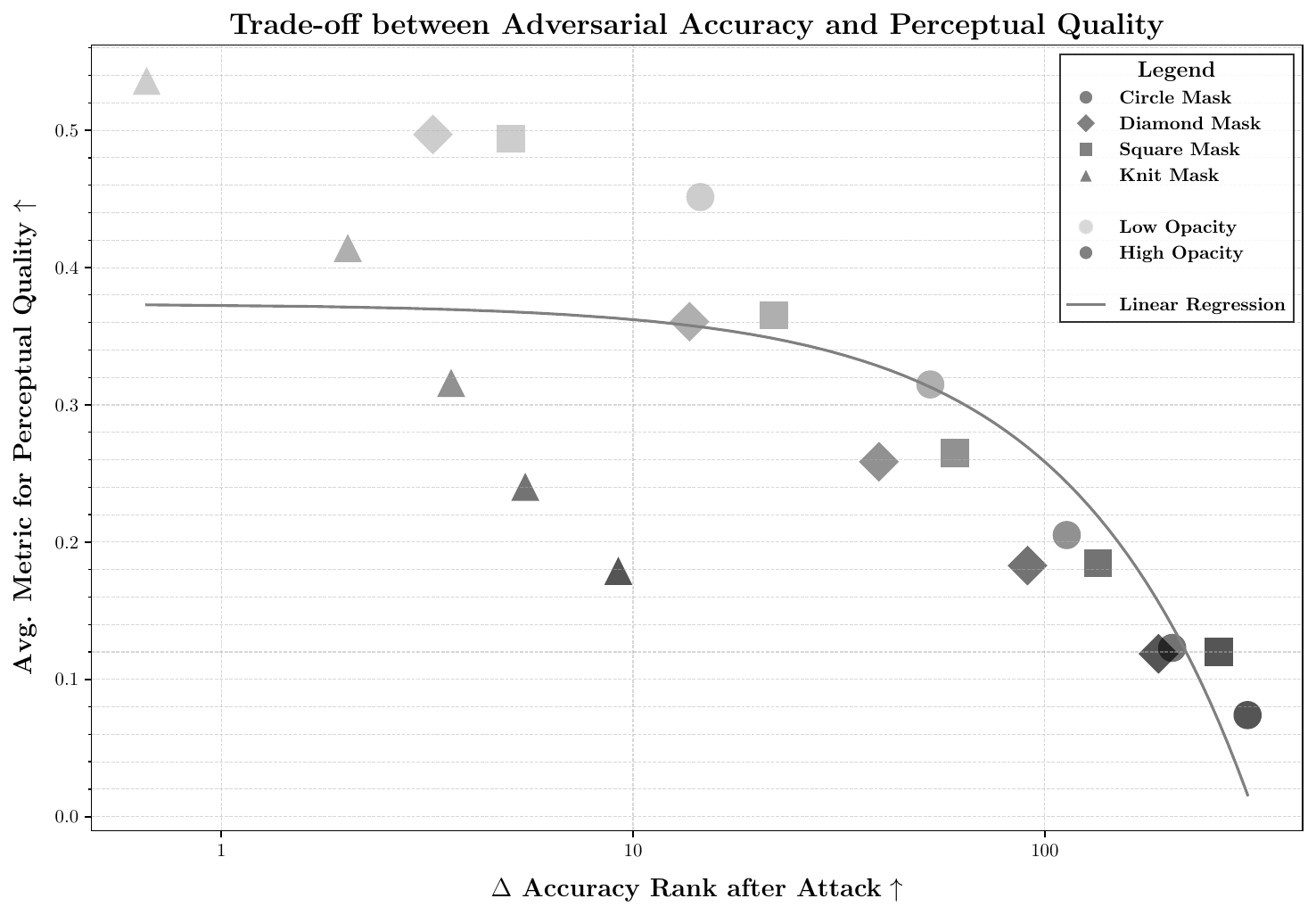}
  \caption{Accuracy vs. Perceptual Quality Trade-off}
  \label{fig:generalizabilityplot}
\end{figure}

We evaluate how the rank of the correct class changes when applying the masks by measuring the rank (the position after sorting) of the ground-truth class before and after applying a mask to an image. In addition, we measure the perceptual quality of the images. We then look at the mean change in rank across models and images, and report the results for each combination of mask and opacity. 

The results of our experiment are visualized in \Cref{fig:generalizabilityplot} (the specific values can be found in \Cref{tab:generalizability} in the Appendix). The figure reveals a clear trend in the trade-off between adversarial effectiveness and perceptual quality. The plot shows a clear inverse relationship between these two factors, as indicated by the polynomial regression curve of degree 2. This relationship suggests that as the effectiveness of the adversarial attack increases (lower $\Delta$ Accuracy Rank), the perceptual quality of adversarial examples tends to decrease. This could be expected, but we noticeably see instances with significant drops in rank (>10) while having a relatively high perceptual quality (>0.4).

The different mask types (circle, square, diamond, and knit) and opacity levels demonstrate varying performance across this trade-off spectrum. The scatter plot reveals clusters of points corresponding to different mask types, with some masks consistently outperforming others in terms of balancing attack effectiveness and perceptual quality. Most importantly, it shows that these geometric pattern masks generalize across SOTA models.

\subsection{Experiment 2 -- \texttt{SubSet200}}

\begin{table}[b]
    \setlength{\tabcolsep}{2pt}
    \footnotesize

    \centering    
    \begin{tabular}{ll|cccc}
        \multicolumn{2}{c|}{} & \multicolumn{4}{c}{Opacity} \\
        Model & Mask & 20\% & 30\% & 40\% & 50\% \\
        \midrule
        \multirow{3}{*}{ConvNeXt} & Circle  & 15.36 (4.40) & 28.49 (12.47) & 43.73 (24.76) & 62.11 (40.72) \\
         & Diamond                          & 3.86 (0.36) & 9.22 (2.11) & 18.55 (6.20) & 34.40 (16.14) \\
         & Square                           & 6.51 (0.90) & 18.73 (5.30) & 35.54 (15.00) & 55.90 (32.53) \\\midrule
        \multirow{3}{*}{EVA02} & Circle     & 10.78 (1.33) & 21.63 (5.60) & 34.22 (14.58) & 43.55 (27.17) \\
         & Diamond                          & 1.87 (0.00) & 6.63 (0.30) & 15.12 (1.81) & 26.33 (5.78) \\
         & Square                           & 6.93 (0.30) & 16.02 (2.23) & 28.73 (8.43) & 41.69 (19.64) \\\midrule
        \multirow{3}{*}{ResNet} & Circle    & 19.70 (5.66) & 32.35 (12.35) & 45.36 (22.47) & 59.94 (33.31) \\
         & Diamond                          & 10.12 (2.23) & 25.30 (10.12) & 47.83 (24.94) & 68.73 (45.72) \\
         & Square                           & 12.65 (3.19) & 27.23 (10.96) & 47.11 (24.76) & 67.65 (41.87) \\\midrule
        \multirow{3}{*}{ViT-H-14} & Circle  & 4.22 (0.78) & 11.75 (3.31) & 27.59 (12.17) & 49.40 (28.25) \\
         & Diamond                          & 0.72 (0.00) & 1.39 (0.24) & 3.07 (0.42) & 7.41 (2.05) \\
         & Square                           & 1.81 (0.06) & 3.25 (0.66) & 12.59 (3.80) & 31.45 (17.17) \\\midrule
        \multirow{3}{*}{RoBERTa-L} & Circle & 7.29 (1.93) & 21.51 (8.31) & 42.77 (21.75) & 62.89 (39.70) \\
         & Diamond                          & 1.51 (0.06) & 4.82 (0.96) & 12.41 (3.25) & 25.12 (9.82) \\
         & Square                           & 4.76 (0.84) & 12.83 (3.07) & 28.73 (11.20) & 52.83 (30.00) \\
        \end{tabular}
        \vspace*{0.1cm}
        \caption{Change of Acc@1 (and Acc@5) for \texttt{SubSet200} [\%].}
        \label{tab:Acc@k for SubSet200}
\end{table}
This experiment measures the drop in Acc@1 and Acc@5 for the subset of images in \texttt{SubSet200} that all models correctly classify. Thus, for the images used in this experiment, Acc@1 (and Acc@5) is 100\% before applying the masks. 
We show in \Cref{tab:Acc@k for SubSet200} the change in accuracy observed in the experiment. The table shows that the circle mask is very effective in confusing models, and even with a relatively low opacity the Acc@1 drops by almost 20\%-points for ResNet. We also see that RoBERTa, as a supposedly robust model, is worse than ViT for masks and opacity levels. Based on the results, we see that diamond-shaped masks pose the least threat to the models at any opacity, but the square masks are almost as effective as the circle masks. In an extension of this, we also looked at the confidence scores, the results of which are \Cref{sec:ground truth resizedall}.

\subsection{Experiment 3 -- \texttt{ResizedAll}}

In this experiment, we used the \texttt{ResizedAll} dataset to measure the drop in Acc@1 and Acc@5 of the models for CAPTCHA-sized images. We see the result of this in \Cref{tab:Acc@k for All}, and an important conclusion regarding the combination of masks and resolution changes is that while the drops in Acc@1 are similar to earlier, the drops in Acc@5 are larger. Compared to the results from the previous experiment, it is evident that in this setting, masks at much lower opacity ratios are more successful in distorting models' performance. Based on these results, the scaling of images combines very well with masks. In closer analysis, it is also evident that EVA02 is the one that suffers the least from circular masks at opacity values >30\% in both datasets, but that it comes at a trade-off of being more sensitive to diamond-shaped masks.

\section{Conclusion}

In this study, we have demonstrated the high effectiveness of geometric masks in fooling state-of-the-art vision models, and the experiments leverage the gaps between human and machine abilities. This suggests potential new directions for developing more robust vision models over the long term while creating secure visual challenges in the short term. 
We show that there is a clear trade-off in the perceptual quality of images for them to be effective against vision models. However, while the perceptual quality decreases, the accuracy of the models also drops, often with more than 50\%-points. This highlights vulnerabilities in advanced vision systems and underscores the continued capability of CAPTCHA-style challenges in differentiating humans from machines.

Although our study focused on specific mask types and datasets, one could easily expand into other masks or determine how effectively models can be fine-tuned on images with masks applied. Furthermore, one could try the methods on the recently published DeepMind model which is supposed to be very robust against adversarial examples~\citep{fort2024ensembleeverywheremultiscaleaggregation}. In addition, a detailed human evaluation of the masks should be performed.

Overall, this study contributes to the ongoing discussion on AI safety and reliability, highlighting the persistent challenge of creating truly robust vision systems that can match human-level adaptability in visual perception tasks.

\begin{table}[t]
    \centering    
    \begin{tabular}{ll|cccc}
        \multicolumn{2}{c|}{} & \multicolumn{4}{c}{Opacity} \\
        Model & Mask & 20\% & 30\% & 40\% & 50\% \\
        \midrule
        \multirow{3}{*}{ConvNeXt} & Circle  & 29.19 (22.42) & 60.03 (54.46) & 77.90 (81.72) & 83.17 (92.15) \\
         & Diamond                          & 14.13 (7.85) & 27.61 (17.67) & 44.90 (34.26) & 60.64 (55.38) \\
         & Square                           & 19.20 (9.64) & 34.41 (22.54) & 56.46 (46.97) & 73.45 (73.76) \\\midrule
        \multirow{3}{*}{EVA02} & Circle     & 31.79 (18.02) & 49.85 (35.62) & 60.88 (51.76) & 70.18 (64.66) \\
         & Diamond                          & 18.53 (5.83) & 30.72 (12.45) & 44.20 (23.03) & 55.31 (36.42) \\
         & Square                           & 23.75 (8.20) & 39.69 (19.66) & 57.61 (40.56) & 69.98 (61.78) \\\midrule
        \multirow{3}{*}{ResNet} & Circle    & 63.53 (48.57) & 76.44 (69.36) & 79.43 (73.21) & 80.14 (74.74) \\
         & Diamond                          & 42.94 (23.41) & 69.33 (50.58) & 82.46 (73.29) & 86.93 (87.20) \\
         & Square                           & 36.27 (19.40) & 66.85 (51.60) & 83.77 (81.84) & 88.41 (94.35) \\\midrule
        \multirow{3}{*}{ViT-H-14} & Circle  & 21.15 (8.89) & 47.78 (26.80) & 71.36 (51.07) & 85.55 (71.71) \\
         & Diamond                          & 5.26 (1.25) & 10.55 (3.33) & 18.80 (8.50) & 32.89 (20.92) \\
         & Square                           & 10.78 (4.17) & 26.94 (15.49) & 55.77 (43.32) & 78.86 (71.37) \\\midrule
        \multirow{3}{*}{RoBERTa-L} & Circle & 37.21 (17.90) & 66.50 (47.53) & 83.84 (73.32) & 91.09 (85.24) \\
         & Diamond                          & 12.64 (3.84) & 24.93 (10.62) & 43.00 (22.82) & 59.68 (40.75) \\
         & Square                           & 19.83 (6.32) & 40.93 (20.68) & 68.47 (52.44) & 86.17 (80.77) \\
    \end{tabular}
    \vspace*{0.1cm}
    \caption{Change of Acc@1 (and Acc@5) for resized ImageNette (\texttt{ResizedAll}) [\%].}
    \label{tab:Acc@k for All}
\end{table}

\appendix

\newpage
\clearpage

\section{Appendix / supplemental material}

\subsection{Acc@1 and Acc@5 accuracy of the tested models.}\label{sec:Acc@1 and Acc@5 All Clean}
\begin{table}[h]
    \setlength{\tabcolsep}{2pt}
    \footnotesize
    
    \centering
    \begin{tabular}{l|cc}
        Model & Acc@1 (\%) & Acc@5 (\%) \\
        \midrule
        ConvNeXt & 84.75 & 95.82 \\
        EVA02 & 92.67 & 97.97 \\
        Apple: ViT-H & 93.10 & 99.29 \\
        ResNet & 89.54 & 98.26 \\
        ViT-H-14 & 93.10 & 99.29 \\
        ViT-L-14 & 91.47 & 98.77 \\
        RoBERTa-B & 84.61 & 97.18 \\
        RoBERTa-L & 93.61 & 98.45 \\
    \end{tabular}
    \vspace*{0.1cm}
    \caption{Acc@1 and Acc@5 (in \%) for Different Models on the \texttt{ResizedAll} dataset.}
    \label{tab:model_accuracy_on_resized_clean}
\end{table}

\subsection{Hyperparameter Optimization}\label{sec:hyperparam}

In our hyperparameter optimization phase, we focused on classification models because of their interpretability advantages over segmentation models. Our initial dataset comprised 1600 scraped and annotated hCaptcha samples, which we used to benchmark several state-of-the-art closed-vocabulary classification models. The ``EVA01-g-14 model'', trained on ``LAION-400M'', emerged as the top performer with Acc@1 of 94.39\% and Acc@5 of 98.93\%. Other models like ``ConvNeXt-XXLarge'' and ``ViT-H-14'' also showed strong performance, although none achieved 100\% accuracy, a notable departure from the results typically seen with reCAPTCHAv2~\citep{20.500.11850/679226}.

Upon analysis of the misclassified images, we observed a combination of imperceptible perturbations and perceptible geometric masks. We identified four distinct geometric mask types for reconstruction and added a novel ``knit'' mask, essentially a modified ``diamond'' mask allowing for overlapping shapes. We intentionally left out word-level adversarial attack masks, as they have been proven to be easy to mitigate~\citep{zhang2023text,dong2023robust,shayegani2023plug}. For each mask, we parameterized three variables: ``opacity'' (alpha value of the overlay), ``density'' (shapes per row/column and nesting, ranging from 0-100), and ``epsilon'' (for white-box FGSM attacks with CLIP-ViT on ImageNet).

We conducted a hyperparameter grid search using the \texttt{visual-layer/imagenet-1k-vl-enriched} dataset on HuggingFace, testing 5-20 examples per combination on the validation set. We chose the CLIP ViT model for this phase due to its superior adversarial robustness, as noted by~\citet{wang2024roz}. Our optimization metric combined the difference in model accuracy pre- and post-mask application with an average of three perceptual quality metrics. To identify optimal parameters, we selected examples with the highest perceptual quality for each level of accuracy difference and performed a linear regression. We then focused on samples above the regression line in multidimensional space. This approach proved to be more tractable than our attempts with multi-objective optimization with multiple variables.

Our findings revealed that FGSM perturbations generally degraded the results when combined with masks. We determined that the optimal density value was consistently 70, while the most effective opacity range was 50-170 (equivalent to 19\%-66\% alpha). These insights allowed us to isolate the best-performing masks for a comprehensive benchmark against the latest models.

This rigorous optimization process, grounded in semantic computer vision research, enabled us to systematically explore the parameter space and identify the most effective adversarial techniques inspired by hCaptcha challenges. The results, visualized in~\Cref{fig:hcaptcha-combined}, provide a quantitative basis for comparing the masks.

\subsection{Generalizability of Masks -- Table}
The table with values plotted in \Cref{fig:generalizabilityplot} can be found in \Cref{tab:generalizability}.

\begin{table}[h]
    \centering
    \begin{tabular}{rlrrr}
        \toprule
        Opacity & Mask & $\Delta$ Acc Rank & Quality & Score \\
        \midrule
        \multirow{4}{*}{50}
            & Circle & -14.57 & 0.45 & 15.02 \\
            & Diamond & -3.27 & 0.50 & 3.76 \\
            & Knit & -0.66 & 0.54 & 1.19 \\
            & Square & -5.04 & 0.49 & 5.54 \\
        \midrule
        \multirow{4}{*}{80}
            & Circle & -52.72 & 0.31 & 53.03 \\
            & Diamond & -13.72 & 0.36 & 14.08 \\
            & Knit & -2.03 & 0.41 & 2.44 \\
            & Square & -22.01 & 0.37 & 22.37 \\
        \midrule
        \multirow{4}{*}{110}
            & Circle & -113.07 & 0.21 & 113.27 \\
            & Diamond & -39.55 & 0.26 & 39.81 \\
            & Knit & -3.62 & 0.32 & 3.93 \\
            & Square & -60.57 & 0.27 & 60.84 \\
        \midrule
        \multirow{4}{*}{140}
            & Circle & -203.89 & 0.12 & 204.01 \\
            & Diamond & -90.79 & 0.18 & 90.97 \\
            & Knit & -5.47 & 0.24 & 5.71 \\
            & Square & -134.75 & 0.18 & 134.94 \\
        \midrule
        \multirow{4}{*}{170}
            & Circle & -310.80 & 0.07 & 310.88 \\
            & Diamond & -188.92 & 0.12 & 189.04 \\
            & Knit & -9.21 & 0.18 & 9.39 \\
            & Square & -264.90 & 0.12 & 265.02 \\
        \bottomrule
    \end{tabular}
    \vspace*{0.1cm}
    \caption{Generalizability of Masks}
    \label{tab:generalizability}
\end{table}

\subsection{Acc@1 and Acc@5 accuracy for \texttt{SubSet500}.}

In \Cref{tab:Drop Acc@1 SubSet500,tab:Drop Acc@5 SubSet500} we show the full tables with drops in accuracy for all the tested models. We see that the circle mask is very aggressive against all models.

\begin{table}[h]
    \setlength{\tabcolsep}{2pt}
    \tiny
    \centering
    \begin{minipage}{0.48\textwidth}
        \centering

        \begin{tabular}{ll|rrrrr}
            \multicolumn{2}{c|}{} & \multicolumn{5}{c}{Opacity} \\
        Model & Mask &  19\% & 31\% & 43\% & 54\% & 66\% \\
        \midrule
        \multirow{4}{*}{ConvNeXt} & Circle & 13.0  & 33.6  & 51.2  & 64.6  & 69.2  \\
         & Diamond & 4.8  & 13.6  & 31.8  & 49.6  & 64.6  \\
         & Knit & 2.2  & 3.2  & 8.0  & 11.4  & 18.0  \\
         & Square & 6.8  & 18.4  & 36.4  & 52.0  & 65.6 \\\midrule
        \multirow{4}{*}{EVA01} & Circle & 7.2  & 15.4  & 33.0  & 49.2  & 65.0  \\
         & Diamond & 2.6  & 8.6  & 19.6  & 33.0  & 54.8  \\
         & Knit & 1.2  & 1.2  & 4.4  & 6.6  & 10.6  \\
         & Square & 4.2  & 9.0  & 17.4  & 31.4  & 55.8 \\\midrule
        \multirow{4}{*}{EVA02} & Circle & 9.4  & 19.0  & 31.4  & 50.4  & 63.8  \\
         & Diamond & 2.4  & 5.6  & 10.6  & 19.0  & 38.0  \\
         & Knit & 2.8  & 4.8  & 5.2  & 6.8  & 8.8  \\
         & Square & 6.8  & 12.4  & 20.8  & 37.4  & 61.8 \\\midrule
        \multirow{4}{*}{ResNet} & Circle & 31.0  & 54.6  & 60.0  & 62.4  & 63.4  \\
         & Diamond & 13.2  & 31.6  & 50.4  & 59.4  & 62.2  \\
         & Knit & 5.0  & 11.2  & 14.4  & 19.4  & 27.6  \\
         & Square & 15.2  & 38.8  & 56.0  & 62.2  & 63.4 \\\midrule
        \multirow{4}{*}{ViT-H-14} & Circle & 5.8  & 20.6  & 48.2  & 70.8  & 80.2  \\
         & Diamond & 2.0  & 5.4  & 15.2  & 34.4  & 61.8  \\
         & Knit & 1.6 & 2.4  & 2.8  & 6.2  & 8.0  \\
         & Square & 3.2  & 9.6  & 25.0  & 54.2  & 77.2  \\
        \end{tabular}
        \vspace*{0.1cm}
        \caption{Change of Acc@1 for \texttt{SubSet500} [\%].}
        \label{tab:Drop Acc@1 SubSet500}
        
    \end{minipage}%
    \hfill
    \begin{minipage}{0.48\textwidth}
        \centering

        \begin{tabular}{ll|rrrrr}
            \multicolumn{2}{c|}{} & \multicolumn{5}{c}{Opacity} \\
            Model & Mask &  19\% & 31\% & 43\% & 54\% & 66\% \\
            \midrule
            \multirow{4}{*}{ConvNeXt} & Circle & 7.60 & 29.60 & 54.80 & 73.40 & 85.00 \\
             & Diamond & 2.60 & 8.80 & 24.20 & 51.40 & 71.60 \\
             & Knit & 1.80 & 2.20 & 4.60 & 7.80 & 13.20 \\
             & Square & 4.80 & 13.20 & 28.80 & 54.80 & 76.80 \\\midrule
            \multirow{4}{*}{EVA01} & Circle & 4.80 & 14.00 & 27.80 & 50.60 & 75.40 \\
             & Diamond & 2.40 & 6.60 & 14.80 & 31.00 & 57.60 \\
             & Knit & 1.40 & 2.80 & 4.60 & 6.20 & 8.00 \\
             & Square & 3.40 & 7.00 & 12.60 & 28.20 & 61.00 \\\midrule
            \multirow{4}{*}{EVA02} & Circle & 4.60 & 12.20 & 24.40 & 44.60 & 65.00 \\
             & Diamond & 1.40 & 3.60 & 6.60 & 14.80 & 34.80 \\
             & Knit & 0.40 & 0.40 & 1.80 & 3.20 & 4.80 \\
             & Square & 2.20 & 6.60 & 15.00 & 31.40 & 63.60 \\\midrule
            \multirow{4}{*}{ResNet} & Circle & 34.20 & 67.40 & 80.40 & 85.40 & 86.20 \\
             & Diamond & 12.20 & 28.80 & 56.00 & 75.60 & 85.00 \\
             & Knit & 4.40 & 8.00 & 10.20 & 15.00 & 20.80 \\
             & Square & 15.20 & 40.20 & 66.40 & 82.20 & 86.60 \\\midrule
            \multirow{4}{*}{ViT-H-14} & Circle & 2.60 & 16.20 & 46.00 & 77.60 & 90.80 \\
             & Diamond & 0.20 & 2.20 & 10.60 & 28.60 & 61.20 \\
             & Knit & -0.60 & 0.60 & 1.00 & 2.00 & 3.20 \\
             & Square & 1.40 & 6.40 & 18.60 & 50.20 & 82.80 
        \end{tabular}
        \vspace*{0.1cm}
        \caption{Change of Acc@5 for \texttt{SubSet500} [\%].}
        \label{tab:Drop Acc@5 SubSet500}

    \end{minipage}
\end{table}

\subsection{Acc@1 and Acc@5 accuracy for \texttt{SubSet200}.}

In the following we show the full tables with Acc@1 and Acc@5 in \Cref{tab:Drop Acc@1 SubSet200,tab:Drop Acc@5 SubSet200} when evaluating on \texttt{SubSet200} as done in Experiment 2. Noticeably, RoBERTa-B performs much worse than RoBERTa-L as its accuracy drops much more. As mentioned in the main results, we see in general that the models have a harder time dealing with the ``circles'' mask. 

\begin{table}[h]
    \setlength{\tabcolsep}{2pt}
    \tiny
    \centering
    \begin{minipage}{0.48\textwidth}
        \centering

        \begin{tabular}{ll|ccccc}
            \multicolumn{2}{c|}{} & \multicolumn{5}{c}{Opacity} \\
            Model & Mask & 10\% & 20\% & 30\% & 40\% & 50\% \\
            \midrule
            \multirow{3}{*}{ConvNeXt} & Circle & 4.46 & 15.36 & 28.49 & 43.73 & 62.11 \\
             & Diamond & 0.78 & 3.86 & 9.22 & 18.55 & 34.40 \\
             & Square & 1.39 & 6.51 & 18.73 & 35.54 & 55.90 \\\midrule
            \multirow{3}{*}{EVA02} & Circle & 1.27 & 10.78 & 21.63 & 34.22 & 43.55 \\
             & Diamond & 0.54 & 1.87 & 6.63 & 15.12 & 26.33 \\
             & Square & 1.20 & 6.93 & 16.02 & 28.73 & 41.69 \\\midrule
            \multirow{3}{*}{Apple: ViT-H} & Circle & 1.02 & 4.22 & 11.75 & 27.59 & 49.40 \\
             & Diamond & 0.36 & 0.72 & 1.39 & 3.07 & 7.41 \\
             & Square & 0.78 & 1.81 & 3.25 & 12.59 & 31.45 \\\midrule
            \multirow{3}{*}{ResNet} & Circle & 5.24 & 19.70 & 32.35 & 45.36 & 59.94 \\
             & Diamond & 2.05 & 10.12 & 25.30 & 47.83 & 68.73 \\
             & Square & 2.89 & 12.65 & 27.23 & 47.11 & 67.65 \\\midrule
            \multirow{3}{*}{ViT-H-14} & Circle & 1.02 & 4.22 & 11.75 & 27.59 & 49.40 \\
             & Diamond & 0.36 & 0.72 & 1.39 & 3.07 & 7.41 \\
             & Square & 0.78 & 1.81 & 3.25 & 12.59 & 31.45 \\\midrule
            \multirow{3}{*}{ViT-L-14} & Circle & 1.93 & 6.93 & 13.67 & 20.42 & 29.88 \\
             & Diamond & 0.30 & 1.33 & 2.59 & 5.84 & 11.69 \\
             & Square & 1.69 & 6.08 & 10.42 & 16.02 & 26.57 \\\midrule
            \multirow{3}{*}{RoBERTa-B} & Circle & 10.84 & 36.81 & 61.51 & 78.31 & 90.12 \\
             & Diamond & 3.13 & 10.06 & 23.67 & 42.23 & 61.14 \\
             & Square & 7.35 & 22.29 & 39.40 & 64.70 & 83.92 \\\midrule
            \multirow{3}{*}{RoBERTa-L} & Circle & 1.02 & 7.29 & 21.51 & 42.77 & 62.89 \\
             & Diamond & 0.42 & 1.51 & 4.82 & 12.41 & 25.12 \\
             & Square & 0.78 & 4.76 & 12.83 & 28.73 & 52.83 \\
        \end{tabular}
        \vspace*{0.1cm}
        \caption{Change of Acc@1 for \texttt{SubSet200} [\%].}
        \label{tab:Drop Acc@1 SubSet200}
    
    \end{minipage}%
    \hfill
    \begin{minipage}{0.48\textwidth}
        \centering
        
        \begin{tabular}{ll|ccccc}
            \multicolumn{2}{c|}{} & \multicolumn{5}{c}{Opacity} \\
            Model & Mask & 10\% & 20\% & 30\% & 40\% & 50\% \\
            \midrule
            \multirow{3}{*}{ConvNeXt} & Circle & 0.48) & 4.40) & 12.47) & 24.76) & 40.72) \\
             & Diamond & 0.00) & 0.36) & 2.11) & 6.20) & 16.14) \\
             & Square & 0.06 & 0.90 & 5.30 & 15.00 & 32.53 \\\midrule
            \multirow{3}{*}{EVA02} & Circle & 0.06 & 1.33 & 5.60 & 14.58 & 27.17 \\
             & Diamond & 0.00 & 0.00 & 0.30 & 1.81 & 5.78 \\
             & Square & 0.00 & 0.30 & 2.23 & 8.43 & 19.64 \\\midrule
            \multirow{3}{*}{Apple: ViT-H} & Circle & 0.06 & 0.78 & 3.31 & 12.17 & 28.25 \\
             & Diamond & 0.00 & 0.00 & 0.24 & 0.42 & 2.05 \\
             & Square & 0.00 & 0.06 & 0.66 & 3.80 & 17.17 \\\midrule
            \multirow{3}{*}{ResNet} & Circle & 0.84 & 5.66 & 12.35 & 22.47 & 33.31 \\
             & Diamond & 0.18 & 2.23 & 10.12 & 24.94 & 45.72 \\
             & Square & 0.42 & 3.19 & 10.96 & 24.76 & 41.87 \\\midrule
            \multirow{3}{*}{ViT-H-14} & Circle & 0.06 & 0.78 & 3.31 & 12.17 & 28.25 \\
             & Diamond & 0.00 & 0.00 & 0.24 & 0.42 & 2.05 \\
             & Square & 0.00 & 0.06 & 0.66 & 3.80 & 17.17 \\\midrule
            \multirow{3}{*}{ViT-L-14} & Circle & 0.12 & 1.45 & 3.98 & 7.89 & 13.43 \\
             & Diamond & 0.00 & 0.12 & 0.42 & 0.96 & 2.65 \\
             & Square & 0.00 & 0.60 & 2.47 & 5.24 & 10.06 \\\midrule
            \multirow{3}{*}{RoBERTa-B} & Circle & 1.57 & 13.19 & 35.72 & 57.29 & 74.46 \\
             & Diamond & 0.00 & 1.39 & 5.96 & 16.45 & 34.64 \\
             & Square & 0.36 & 4.82 & 13.73 & 37.11 & 65.54 \\\midrule
            \multirow{3}{*}{RoBERTa-L} & Circle & 0.06 & 1.93 & 8.31 & 21.75 & 39.70 \\
             & Diamond & 0.00 & 0.06 & 0.96 & 3.25 & 9.82 \\
             & Square & 0.00 & 0.84 & 3.07 & 11.20 & 30.00 \\
        \end{tabular}
        \vspace*{0.1cm}
        \caption{Change of Acc@5 for \texttt{SubSet200} [\%].}
        \label{tab:Drop Acc@5 SubSet200}
    
    \end{minipage}
\end{table}

\subsection{Acc@1 and Acc@5 accuracy for \texttt{ResizedAll}.}

\Cref{tab:Drop Acc@1 All} and \Cref{tab:Drop Acc@5 All} show the full tables with the drops in Acc@1 and Acc@5, respectively when applying the masks to the resized images in ImageNette (\texttt{ResizedAll}).

\begin{table}[h]
    \setlength{\tabcolsep}{2pt}
    \tiny
    \centering
    \begin{minipage}{0.48\textwidth}
        \centering

        \begin{tabular}{ll|ccccc}
            \multicolumn{2}{c|}{} & \multicolumn{5}{c}{Opacity} \\
            Model & Mask & 10\% & 20\% & 30\% & 40\% & 50\% \\
            \midrule
            \multirow{3}{*}{ConvNeXt} & Circle & 5.59 & 29.19 & 60.03 & 77.90 & 83.17 \\
             & Diamond & 2.14 & 14.13 & 27.61 & 44.90 & 60.64 \\
             & Square & 5.95 & 19.20 & 34.41 & 56.46 & 73.45 \\\midrule
            \multirow{3}{*}{EVA02} & Circle & 5.52 & 31.79 & 49.85 & 60.88 & 70.18 \\
             & Diamond & 6.32 & 18.53 & 30.72 & 44.20 & 55.31 \\
             & Square & 13.79 & 23.75 & 39.69 & 57.61 & 69.98 \\\midrule
            \multirow{3}{*}{Apple: ViT-H} & Circle & 2.80 & 21.15 & 47.78 & 71.36 & 85.55 \\
             & Diamond & -0.14 & 5.26 & 10.55 & 18.80 & 32.89 \\
             & Square & 2.34 & 10.78 & 26.94 & 55.77 & 78.86 \\\midrule
            \multirow{3}{*}{ResNet} & Circle & 17.02 & 63.53 & 76.44 & 79.43 & 80.14 \\
             & Diamond & 10.86 & 42.94 & 69.33 & 82.46 & 86.93 \\
             & Square & 10.28 & 36.27 & 66.85 & 83.77 & 88.41 \\\midrule
            \multirow{3}{*}{ViT-H-14} & Circle & 2.80 & 21.15 & 47.78 & 71.36 & 85.55 \\
             & Diamond & -0.14 & 5.26 & 10.55 & 18.80 & 32.89 \\
             & Square & 2.34 & 10.78 & 26.94 & 55.77 & 78.86 \\\midrule
            \multirow{3}{*}{ViT-L-14} & Circle & 9.17 & 28.73 & 44.13 & 57.61 & 67.10 \\
             & Diamond & 3.20 & 9.58 & 17.10 & 28.60 & 42.83 \\
             & Square & 6.22 & 15.86 & 27.39 & 43.12 & 60.12 \\\midrule
            \multirow{3}{*}{RoBERTa-B} & Circle & 12.68 & 43.45 & 65.91 & 80.44 & 84.31 \\
             & Diamond & 5.55 & 22.64 & 39.30 & 56.07 & 70.51 \\
             & Square & 10.11 & 29.60 & 52.57 & 73.37 & 82.81 \\\midrule
            \multirow{3}{*}{RoBERTa-L} & Circle & 7.69 & 37.21 & 66.50 & 83.84 & 91.09 \\
             & Diamond & 4.30 & 12.64 & 24.93 & 43.00 & 59.68 \\
             & Square & 7.03 & 19.83 & 40.93 & 68.47 & 86.17 \\
        \end{tabular}
        \vspace*{0.1cm}
        \caption{Drop of Acc@1 for resized ImageNette (\texttt{ResizedAll}) [\%].}
        \label{tab:Drop Acc@1 All}
        
    \end{minipage}%
    \hfill
    \begin{minipage}{0.48\textwidth}
        \centering
        
        \begin{tabular}{ll|ccccc}
            \multicolumn{2}{c|}{} & \multicolumn{5}{c}{Opacity} \\
            Model & Mask & 10\% & 20\% & 30\% & 40\% & 50\% \\
            \midrule
            \multirow{3}{*}{ConvNeXt} & Circle & 3.91 & 22.42 & 54.46 & 81.72 & 92.15 \\
             & Diamond & 1.36 & 7.85 & 17.67 & 34.26 & 55.38 \\
             & Square & 2.93 & 9.64 & 22.54 & 46.97 & 73.76 \\\midrule
            \multirow{3}{*}{EVA02} & Circle & 2.53 & 18.02 & 35.62 & 51.76 & 64.66 \\
             & Diamond & 1.85 & 5.83 & 12.45 & 23.03 & 36.42 \\
             & Square & 3.46 & 8.20 & 19.66 & 40.56 & 61.78 \\\midrule
            \multirow{3}{*}{Apple: ViT-H} & Circle & 1.16 & 8.89 & 26.80 & 51.07 & 71.71 \\
             & Diamond & -0.12 & 1.25 & 3.33 & 8.50 & 20.92 \\
             & Square & 0.65 & 4.17 & 15.49 & 43.32 & 71.37 \\\midrule
            \multirow{3}{*}{ResNet} & Circle & 6.64 & 48.57 & 69.36 & 73.21 & 74.74 \\
             & Diamond & 3.54 & 23.41 & 50.58 & 73.29 & 87.20 \\
             & Square & 3.54 & 19.40 & 51.60 & 81.84 & 94.35 \\\midrule
            \multirow{3}{*}{ViT-H-14} & Circle & 1.16 & 8.89 & 26.80 & 51.07 & 71.71 \\
             & Diamond & -0.12 & 1.25 & 3.33 & 8.50 & 20.92 \\
             & Square & 0.65 & 4.17 & 15.49 & 43.32 & 71.37 \\\midrule
            \multirow{3}{*}{ViT-L-14} & Circle & 3.37 & 15.84 & 27.93 & 41.92 & 53.82 \\
             & Diamond & 0.94 & 2.96 & 6.27 & 12.43 & 24.19 \\
             & Square & 2.64 & 6.73 & 12.49 & 24.04 & 42.20 \\\midrule
            \multirow{3}{*}{RoBERTa-B} & Circle & 5.85 & 28.52 & 53.14 & 76.14 & 84.35 \\
             & Diamond & 2.03 & 10.01 & 22.89 & 40.94 & 62.95 \\
             & Square & 3.42 & 15.13 & 36.91 & 67.46 & 87.65 \\\midrule
            \multirow{3}{*}{RoBERTa-L} & Circle & 2.06 & 17.90 & 47.53 & 73.32 & 85.24 \\
             & Diamond & 0.75 & 3.84 & 10.62 & 22.82 & 40.75 \\
             & Square & 1.29 & 6.32 & 20.68 & 52.44 & 80.77 \\
        \end{tabular}
        \vspace*{0.1cm}
        \caption{Drop in Acc@5 for resized ImageNette (\texttt{ResizedAll}) [\%].}
        \label{tab:Drop Acc@5 All}

    \end{minipage}
\end{table}

\subsection{Ground Truth Confidence for \texttt{SubSet200}}\label{sec:ground truth subset200}
In extension to Acc@1 and Acc@5 results, then it is useful to compare the results on the confidence of the ground truth for all the same masks, cf. \Cref{tab:gt_drop_SubSet200}, as it provides a better idea of how stable the Acc@5 scores are. Initially, the confidence in ground truth is very high and stands far from the next prediction for most of the cases shift of up 67\% in confidence for an opacity level in the range of 50\% is not sufficient to drop the model's Acc@5 below 50\%, as it does for Acc@1 and becomes harder for human perception. A drop of ground truth confidence also agrees with the fact that better resistance against some shapes comes at the cost of being more sensitive to the other ones, as it happens based on examples of EVA02 and ViT-H-14.

\subsection{Ground Truth Confidence for \texttt{ResizedAll}}\label{sec:ground truth resizedall}
We see that the confidence of the ground truth drops further for many instances \Cref{tab:gt_drop_resizedAll} which indicates that it can be easier to be combined with an FGSM-like attack and target Acc@5 specifically. The table also demonstrates that VIT-L-14 (not presented in the main sections of the paper) is more resistant to masks of circular shape than ViT-H-14 at opacity levels >40\%, but more sensitive to the other shapes in both datasets.

\begin{table}[h]
    \setlength{\tabcolsep}{2pt}
    \tiny
    \centering
    \begin{minipage}{0.48\textwidth}
        \centering
        \begin{tabular}{ll|ccccc}
            \multicolumn{2}{c|}{} & \multicolumn{5}{c}{Opacity} \\
            Model & Mask & 10\% & 20\% & 30\% & 40\% & 50\% \\
            \midrule
            \multirow{3}{*}{ConvNeXt} & Circles & 5.645 & 18.905 & 33.361 & 49.333 & 66.130 \\
             & Diamond & 1.044 & 4.936 & 11.614 & 22.641 & 38.619 \\
             & Square & 2.067 & 9.481 & 22.784 & 39.798 & 59.099 \\
            \midrule
            \multirow{3}{*}{EVA02} & Circles & 1.589 & 12.123 & 24.320 & 36.740 & 47.974 \\
             & Diamond & -0.169 & 2.543 & 8.605 & 18.073 & 29.397 \\
             & Square & 1.688 & 9.081 & 19.313 & 31.203 & 44.633 \\
            \midrule
            \multirow{3}{*}{Apple: ViT-H} & Circles & 0.041 & 3.274 & 11.708 & 28.551 & 50.050 \\
             & Diamond & -0.545 & -0.587 & 0.059 & 2.073 & 6.705 \\
             & Square & -0.496 & 0.457 & 2.806 & 11.959 & 31.869 \\
            \midrule
            \multirow{3}{*}{ResNet} & Circles & 8.323 & 26.143 & 41.388 & 56.402 & 69.312 \\
             & Diamond & 3.928 & 15.906 & 33.133 & 55.247 & 72.711 \\
             & Square & 4.559 & 17.200 & 33.441 & 53.504 & 71.878 \\
            \midrule
            \multirow{3}{*}{ViT-H-14} & Circles & 0.041 & 3.274 & 11.708 & 28.551 & 50.050 \\
             & Diamond & -0.545 & -0.587 & 0.059 & 2.073 & 6.705 \\
             & Square & -0.496 & 0.457 & 2.806 & 11.959 & 31.869 \\
            \midrule
            \multirow{3}{*}{ViT-L-14} & Circles & 5.912 & 14.392 & 22.576 & 31.488 & 42.215 \\
             & Diamond & 1.229 & 4.030 & 7.882 & 13.611 & 21.541 \\
             & Square & 5.112 & 13.642 & 21.162 & 29.442 & 40.269 \\
            \midrule
            \multirow{3}{*}{RoBERTa-B} & Circles & 10.594 & 37.631 & 60.838 & 75.623 & 85.685 \\
             & Diamond & 2.048 & 10.495 & 24.429 & 43.469 & 61.539 \\
             & Square & 6.831 & 22.112 & 41.473 & 63.348 & 80.237 \\
            \midrule
            \multirow{3}{*}{RoBERTa-L} & Circles & 1.809 & 10.122 & 25.972 & 47.397 & 66.992 \\
             & Diamond & 0.731 & 2.408 & 6.760 & 15.256 & 29.578 \\
             & Square & 1.115 & 6.424 & 15.523 & 32.914 & 56.083 \\
        \end{tabular}
        \vspace*{0.1cm}
        \caption{Ground truth conf drop for \texttt{SubSet200} [\%].}
        \label{tab:gt_drop_SubSet200}
    
    \end{minipage}%
    \hfill
    \begin{minipage}{0.48\textwidth}
        \centering
        
        \begin{tabular}{ll|ccccc}
            \multicolumn{2}{c|}{} & \multicolumn{5}{c}{Opacity} \\
            Model & Mask & 10\% & 20\% & 30\% & 40\% & 50\% \\
            \midrule
            \multirow{3}{*}{ConvNeXt} & Circles & 9.617 & 36.234 & 68.709 & 86.811 & 91.899 \\
             & Diamond & 6.480 & 18.077 & 33.547 & 53.098 & 70.053 \\
             & Square & 10.775 & 23.225 & 40.300 & 64.181 & 82.371 \\
            \midrule
            \multirow{3}{*}{EVA02} & Circles & 6.761 & 33.178 & 53.924 & 67.718 & 78.523 \\
             & Diamond & 7.933 & 18.738 & 31.381 & 44.827 & 56.812 \\
             & Square & 15.902 & 27.058 & 42.733 & 61.524 & 75.634 \\
            \midrule
            \multirow{3}{*}{Apple: ViT-H} & Circles & 3.520 & 21.009 & 48.194 & 73.760 & 88.807 \\
             & Diamond & 1.688 & 5.023 & 9.470 & 17.993 & 33.330 \\
             & Square & 2.825 & 9.920 & 26.234 & 56.025 & 80.729 \\
            \midrule
            \multirow{3}{*}{ResNet} & Circles & 24.530 & 76.365 & 88.196 & 89.980 & 90.633 \\
             & Diamond & 15.953 & 52.373 & 76.141 & 86.627 & 90.562 \\
             & Square & 14.456 & 44.220 & 73.281 & 88.062 & 91.925 \\
            \midrule
            \multirow{3}{*}{ViT-H-14} & Circles & 3.435 & 20.802 & 48.019 & 73.709 & 88.582 \\
             & Diamond & 1.829 & 5.373 & 9.629 & 17.986 & 34.228 \\
             & Square & 4.098 & 10.902 & 27.179 & 56.860 & 81.146 \\
            \midrule
            \multirow{3}{*}{ViT-L-14} & Circles & 14.812 & 36.936 & 53.223 & 66.353 & 75.476 \\
             & Diamond & 9.480 & 16.366 & 24.668 & 36.806 & 50.707 \\
             & Square & 10.737 & 23.081 & 35.193 & 51.176 & 67.486 \\
            \midrule
            \multirow{3}{*}{RoBERTa-B} & Circles & 19.521 & 53.359 & 75.171 & 89.228 & 92.515 \\
             & Diamond & 10.482 & 28.408 & 47.659 & 66.300 & 80.111 \\
             & Square & 16.796 & 38.781 & 63.461 & 82.539 & 91.357 \\
            \midrule
            \multirow{3}{*}{RoBERTa-L} & Circles & 11.479 & 42.415 & 71.558 & 89.207 & 96.184 \\
             & Diamond & 7.859 & 15.928 & 29.738 & 48.025 & 65.297 \\
             & Square & 10.725 & 23.512 & 45.274 & 73.033 & 91.098 \\
        \end{tabular}
        \vspace*{0.1cm}
        \caption{Ground truth conf drop for \texttt{ResizedAll} [\%].}
        \label{tab:gt_drop_resizedAll}
    
    \end{minipage}
\end{table}

\clearpage

\bibliography{main}

\end{document}